\documentclass[letterpaper]{article}
\usepackage{aaai2026}
\usepackage{times}
\usepackage{helvet}
\usepackage{courier}
\usepackage[hyphens]{url}
\usepackage{graphicx}
\usepackage{natbib}
\usepackage{caption}
\usepackage{multirow}
\usepackage{amsmath}
\usepackage{amssymb}
\usepackage{algorithm}
\usepackage{algorithmic}
\usepackage{booktabs}
\usepackage{amssymb}  
\usepackage{pifont}   
\usepackage[utf8]{inputenc}
\usepackage{float}

\title{Domain Generalized Stereo Matching with Uncertainty-guided Data Augmentation}
\author{
Shuangli Du\textsuperscript{\rm 1},
Jing Wang\textsuperscript{\rm 1},
Minghua Zhao\textsuperscript{\rm 1}\thanks{Corresponding author},
Zhenyu Xu\textsuperscript{\rm 2},
Jie Li\textsuperscript{\rm 3}
}

\affiliations{
\textsuperscript{\rm 1}School of Computer Science and Engineering, Xi’an University of Technology, Xi’an, China \\
\textsuperscript{\rm 2}College of Computer Science, Sichuan University, Chengdu, China \\
\textsuperscript{\rm 3}College of Information Science, Shanxi University of Finance and Economics, Taiyuan, Shanxi, China \\
dusl@xaut.edu.cn, 2231221100@stu.xaut.edu.cn, zhaominghua@xaut.edu.cn, sanxu@scu.edu.cn, lijie@sxufe.edu.cn
}

\begin{document}

\maketitle

\begin{abstract}
State-of-the-art stereo matching (SM) models trained on synthetic data often fail to generalize to real data domains due to domain differences, such as color, illumination, contrast, and texture. To address this challenge, we leverage data augmentation to expand the training domain, encouraging the model to acquire robust cross-domain feature representations instead of domain-dependent shortcuts. This paper proposes an uncertainty-guided data augmentation (UgDA) method, which argues that the image statistics in RGB space (mean and standard deviation) carry the domain characteristics. Thus, samples in unseen domains can be generated by properly perturbing these statistics. Furthermore, to simulate more potential domains, Gaussian distributions founded on batch-level statistics are poposed to model the unceratinty of perturbation direction and intensity. Additionally, we further enforce feature consistency between original and augmented data for the same scene, encouraging the model to learn structure aware, shortcuts-invariant feature representations. Our approach is simple, architecture-agnostic, and can be integrated into any SM networks. Extensive experiments on several challenging benchmarks have demonstrated that our method can significantly improve the generalization performance of existing SM networks.
\end{abstract}

\section{Introduction}

Stereo Matching (SM) is a fundamental problem in computer vision, robotics, and autonomous driving. Given a rectified image pair, its objective is to search for corresponding points and calculate dense disparity for 3D reconstruction. Traditional matching methods typically consist of four steps ~\cite{hamid2022stereo,fathy2018hierarchical}: feature extraction, matching cost generation, cost aggregation, disparity calculation, and refinement. Recent works have integrated these steps into end-to-end deep neural networks, achieving impressive accuracy on several datasets or benchmarks ~\cite{yang2019hierarchical,zheng2025diffuvolume}.

However, due to the scarcity of labeled realistic training data, state-of-the-art SM networks are often trained on synthetic data such as SceneFlow ~\cite{mayer2016large}, which fail to generalize  to unseen realistic domains. There are two solutions to this issue: domain adaptation techniques ~\cite{liu2020stereogan,poggi2019guided} and domain generalized approaches ~\cite{chang2023domain,wu2024few}. In this work, we tackle the challenging task of single-domain generalization for SM, where only synthetic data is available for training. This task is more challenging than domain adaptation, which usually require samples from the target environments for adaptation.

\begin{figure*}
    \centering
    \includegraphics[scale=0.72,keepaspectratio]{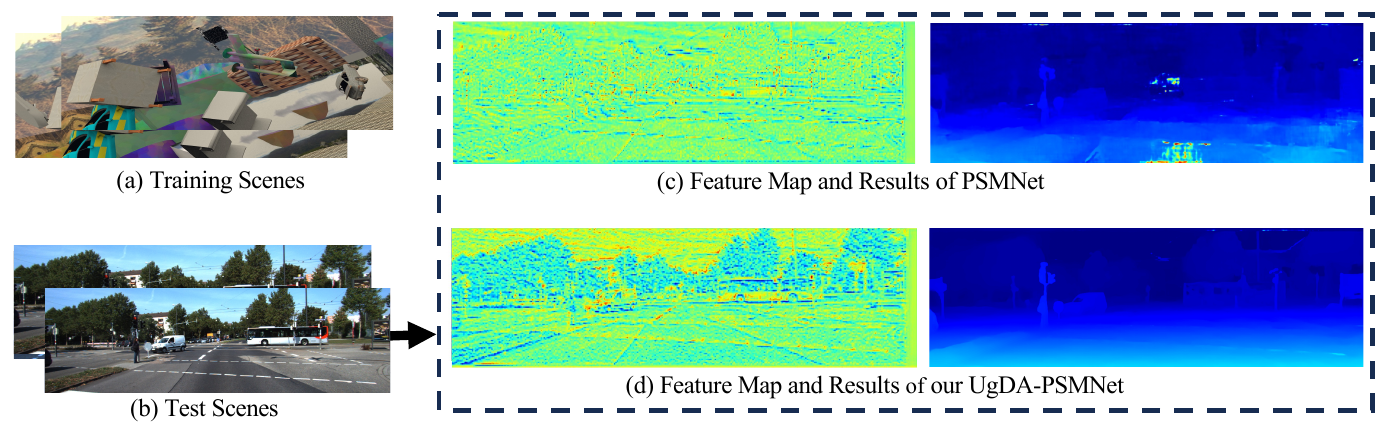}
    \caption{Visualization of the feature maps and disparity results. The PSMNet is used for comparisons. Models are trained on synthetic data (SceneFlow) and tested on novel real scenes (KITTI). The feature maps from PSMNet have many artifacts (e.g., noises). Our UgDA-PSMNet can capture meaningful semantic and structural features, having no distortions or artifacts, and it can obtain  accurate disparity estimations.}
    \label{fig:feature_maps}
\end{figure*}

One key challenge in designing generalizable deep SM models lies in robust feature extraction. This difficulty stems from significant domain shifts between the training data (source domain) and the test data (target domain), particularly in visual characteristics such as color, illumination, contrast, and scene content. When trained solely on limited source-domain data, stereo networks tend to rely on domain-specific shortcuts rather than learning shortcuts-invariant representations. As a result, they fail to extract meaningful semantic and structural features in unseen target domains, as illustrated in Figure~\ref{fig:feature_maps}.

To address this issue, some researchers have attempted to expand training data diversity, using new generated data to learn robust cross-domain feature extractors. For instance, Chang et al.~\cite{chang2023domain} designed multi-level data augmentation modules operating at global, local, and pixel levels. Chuah et al.~\cite{chuah2022itsa} introduced gradient-based perturbations in directions that most affect matching performance, producing adversarial examples. However, how to systematically widening the sample distribution in data augmentation while ensuring diversity and effectiveness remains a major challenge.

In this work, we propose a novel domain generalization strategy for SM based on Uncertainty-guided Data Augmentation (UgDA-stereo), which leverages domain differences aware metric to guide data augmentation. We observe that image RGB channel-wise statistics (i.e., mean and variance) carry the domain characteristics, and can serve as metrics of domain differences. Thus, we propose to model domain shift by directly perturbing these statistics, generating new stylized samples in unseen domains. In addition, the test domains may introduce uncertain domain shifts with varying directions and intensities compared to the training domain. To simulate more potential domains, we explicitly model the uncertainty in both perturbation direction and strength with Gaussian distribution. Specifcally, we quantify the variations of these statistics within a batch, and use the variations to construct Gaussian distribution to describe perturbation uncertainty. The augmented data exhibit diversified visual styles. We further enforce feature consistency between original and augmented data for the same scene, encouraging the network to learn structure aware, shortcuts-invariant feature representations across domains.

Our contributions are summarized as follows:
\begin{itemize}
\item We propose a plug-and-play domain generalization strategy for SM, using uncertainty-driven data augmentation to expand the training domain and force the model to learn robust feature representations. This module is lightweight and requires low computational resources.
\item We introduce a novel uncertainty-guided data augmentation mechanism that perturbs per-image RGB statistics (i.e., mean and variance) to simulate realistic domain shifts. The uncertainty, including perturbation direction and strength is modeled with Gaussian distribution based on batch-level statistical variation. The module is lightweight, architecture-agnostic, and can be integrated into many vision tasks.
\item Extensive experiments on KITTI2012, KITTI2015, Middlebury,  and ETH3D demonstrate that our proposed UgDA-stereo significantly improves the generalization performance of various baseline networks and achieves robust stereo correspondence under challenging weather conditions.
\end{itemize}

\section{Related Work}
\subsection{Deep SM Networks}

In recent years, deep SM models have replaced traditional methods and have excelled in most datasets and benchmarks. MC-CNN~\cite{vzbontar2016stereo} is the first work applying CNN to SM, using CNN to extract matching features while still relying on traditional cost aggregation and disparity computation.  Later, DispNetC ~\cite{mayer2016large} firstly embedded all the core steps of SM into an end-to-end deep network, where the disparity is regressed from the correlation maps through 2D convolutions. Since then, this pipeline has been widely adopted with many SM methods proposed. 

Based on the cost volume construction, these methods can be divided into correlation-based and concatenation-based approaches. Correlation-based methods (e.g., SegStereo ~\cite{yang2018segstereo} and EdgeStereo ~\cite{song2020edgestereo}) compute cross-view feature correlations to build the cost volume, enabling efficient 2D convolution-based disparity estimation. However, they suffer from loss of semantic and structural information in feature representation due to the correlation operation, resulting in limited performance.  Concatenation-based approaches build the cost volume by concatenating the left and the right features and aggregate the cost with 3D convolutions, such as the state-of-the-art PSMNet ~\cite{chang2018pyramid} and GwcNet ~\cite{guo2019group}. For more effective cost aggregation, GA-Net ~\cite{zhang2019ga} introduced content-aware aggregation layers. While theses methods can achieve superior performance with 3D-CNN, they fail to generalize to unseen environments without fine-tuning due to domain shift between the test and train domain.

\subsection{Domain Adaptation and Generalization for SM}

To bridge the gap between synthetic and real-world domains, SM research has explored two key directions: domain adaptation and domain generalization. Domain adaptation methods typically assume access to  target-domain data during training. Representative works include image-level translation strategies like StereoGAN ~\cite{liu2020stereogan}, SDA ~\cite{li2021synthetic}, and Zoom-and-Learn ~\cite{pang2018zoom}, which aim to align cross-domain data while preserving geometric structure. However, these approaches rely on target-domain supervision and require complex optimization or fine-tuning,\begin{figure*}
    \centering
    \includegraphics[scale=0.67,keepaspectratio]{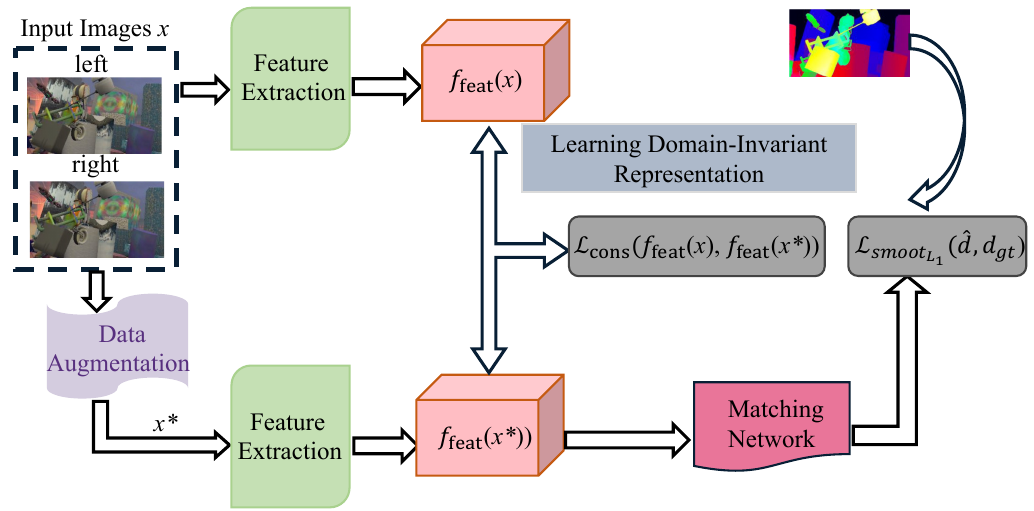}
    \caption{An overview of the proposed  strategy to achieve domain generalization in stereo matching networks. The parameters are shared across the two feature extractor networks.}
    \label{fig:overview}
\end{figure*} making them less practical in deployment.

Unlike adaptation-based methods, domain generalization approaches directly learn SM networks that generalize to unseen domains. RAFT-Stereo employed multi-scale cost volume fusion to extract robust semantic and structural features. Some works propose to learn domain-invariant representations. For example, DSMNet~\cite {zhang2020domain} introduced domain normalization to stabilize features across domains. GraftNet~\cite{liu2022graftnet} leveraged the broad-spectrum feature trained on large-scale datasets to deal with the domain shift since it has seen various styles of images. ITSA~\cite{chuah2022itsa} employed an information-theoretic strategy to minimize feature sensitivity to input perturbations, effectively mitigating shortcut learning. HVT~\cite{chang2023domain} proposed a hierarchical data augmentation method to enforce domain-invariant feature extraction, including global, local, and pixel-wise levels.

Differently, Masked-stereo~\cite{rao2023masked} formulated a pseudo-multi-task learning framework to train SM and image reconstruction jointly, promoting models to learn structure information and to improve generalization performance. Recent innovations have extended these approaches through auxiliary supervision. HODC~\cite{miao2024hierarchical} enforced object- and pixel-level consistency using dual-level contrastive learning, improving robustness to scene-level changes. Meanwhile, Yao et al.~\cite{yao2025diving} fused monocular depth priors through an asymmetric distillation framework, helping suppress stereo ambiguity in textureless or ambiguous regions.

Although these strategies significantly enhance robustness, they often require non-trivial architectural redesigns, extra modalities, or complex learning objectives. This motivates simpler but effective alternatives that can generalize well without modifying backbone structures.

In this paper, we propose a simple plug-and-play domain generalization strategy for SM, using uncertainty-guided data augmentation to expand the training domain and force the model to learn robust feature representations. Our augmentation operates solely at the input level, requires no architectural changes, and explicitly maintains left-right consistency—making it ideal for geometry-sensitive tasks.

\section{Method}

\subsection{Problem Definition}

In this paper, we aim at solving the problem of single domain generalization for SM. Given a source domain with \(N\) labeled samples \(\mathcal{D}_s = \left\{ (x_{Li}, x_{Ri}, d_{gti}) \right\}_{i=1}^N \sim P_s(x_L, x_R, d)\), where \(x_{Li}, x_{Ri} \in \mathbb{R}^{H \times W \times 3}\) are the left and right stereo images, and \(d_{gti} \in \mathbb{R}^{H \times W}\) is the labeled ground-truth disparity, and the source domain’s joint distribution is denoted as \( P_s\),  then a SM network  \(\mathit{F}_\Theta(\cdot,\cdot)\) can be trained with the following formulation:

\begin{equation}
\begin{cases}
\hat{d} = F_\Theta(x_L, x_R) = f_{\text{disp}}(g(f_{\text{feat}}(x_L), f_{\text{feat}}(x_R))) \\
\underset{\Theta}{\min}  \mathbb{E}_{(x_L, x_R, d) \sim P_s} \bigl[ \mathcal{L}(\hat{d}, d_{gt}) \bigr]
\end{cases}
\end{equation}where $f_{\text{feat}}$ denotes  feature extractor, $g(\cdot,\cdot)$ is the cost volume construction module, and $f_{\text{disp}}(\cdot,\cdot)$ is the cost aggregation and disparity regression module. $\hat{d}$ denotes the estimated disparity map, and $d_{gt}$ is the ground-truth disparity. $\mathcal{L}(\cdot,\cdot)$ is the loss function.

Most existing SM models treat disparity estimation as a regression problem, optimizing networks through disparity supervision loss combined with smoothness regularization and left-right consistency constraints. However, such models often suffer from shortcut learning and over-fitting to the training domain. Specifically, the modules $f_{\text{feat}}$, $g$, and $f_{\text{disp}}$ become tightly coupled to the data distribution in domain $\mathcal{D}_s$, and perform poorly on unseen target domain
 \(\mathcal{D}_T = \left\{ (x_{Lj}, x_{Rj}) \right\}_{j=1}^M \sim P_T(x_L, x_R)\) with a different joint distribution $P_T$. And $M$ denotes the number\begin{figure*}
    \centering
    \includegraphics[scale=0.7,keepaspectratio]{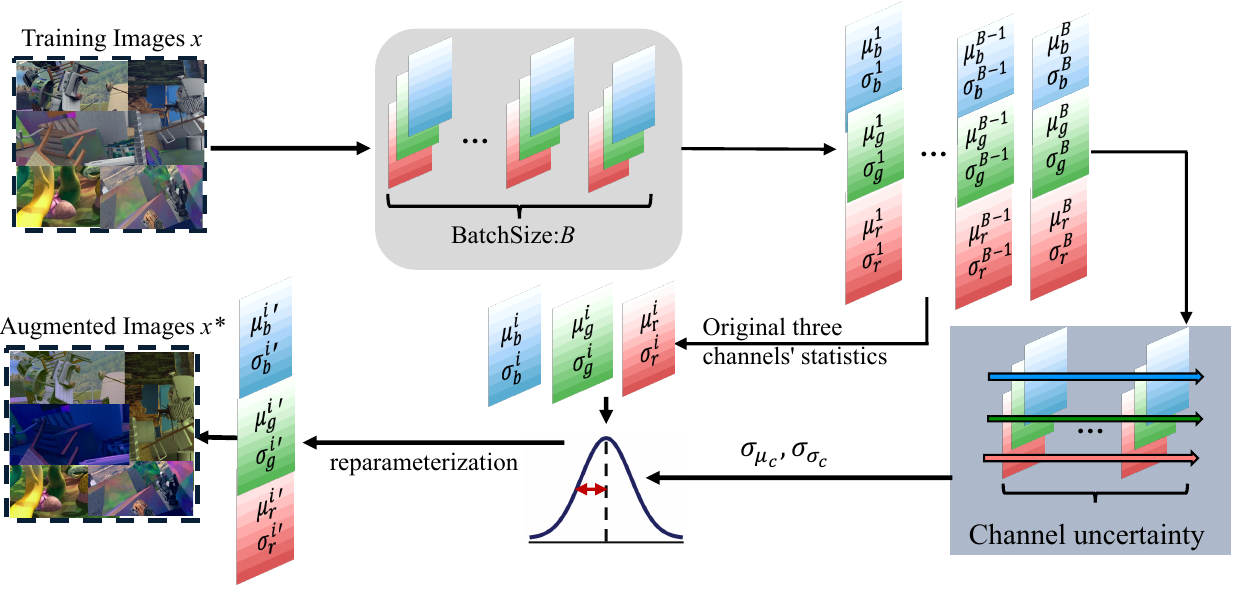}
    \caption{The proposed data augmentation strategy.}
    \label{fig:stylized part}
\end{figure*} of samples. The primary reason is the domain shift between the target and source domains ($P_S \neq P_T$), such as differences in lighting, scene content, and contrast.

Our goal is to train a model on $D_S$ that generalizes well to unseen domains $D_T$ without accessing any target domain samples during training. To achieve this, we develop a data augmentation strategy that generates samples approximating unseen domain distributions to enhance model generalization. The overall workflow of our method is illustrated in Figure~\ref{fig:overview}. An uncertainty-driven data augmentation technique is proposed, which maintains the original scene contents while altering imaging-related properties such as chromatic characteristics, luminance levels, and contrast, etc. Additionally, the approach aims to extract domain-invariant features that capture scene semantics and structure by enforcing feature-level consistency between original and augmented samples. The entire algorithm can be modeled as:


\begin{equation}
\begin{cases}
x_{L}^{*}=A(x_{L}),x_{R}^{*}=A(x_{R}) \\
\hat{d} = F_\Theta(x_{L}^{*}, x_{R}^{*}) = f_{\text{disp}}(g(f_{\text{feat}}(x_{L}^{*}), f_{\text{feat}}(x_{R}^{*})))\\
\underset{\Theta}{\min} \mathbb{E}_{(x_{L}^{*}, x_{R}^{*}, d) \sim P_{s}^*} \bigl[ \mathcal{L}(\hat{d}, d_{gt})+\mathcal{L}_{cons}(f_{\text{feat}}(x),f_{\text{feat}}(x^{*})) \bigr]
\end{cases}
\end{equation}
where \( x_L^*, x_R^* \) denote the augmented left and right images generated from the original stereo pair \( (x_L, x_R) \) via augmentation module \( A \). For augmentation, we sample an image \( x \), either a left or right image, from the unpaired set \(\left\{ x_{Li}, x_{Ri} \mid i=1, \dots, N \right\}\), and  \( x^*\) is the augmented data. The augmented domain $\mathcal{D}_s^*$ follows the distribution $P_s^*(\mathbf{x}_L^*, \mathbf{x}_R^*, \mathit{d})$. \(\mathcal{L}_{\text{cons}}\) is the feature consistency loss. Thus, the trained model has seen both the source domain and the augmented domains. The algorithm will be introduced in detail below.

\subsection{Data Augmentation in RGB Space}

\begin{figure}[t]
\centering
\includegraphics[scale=0.35,keepaspectratio]{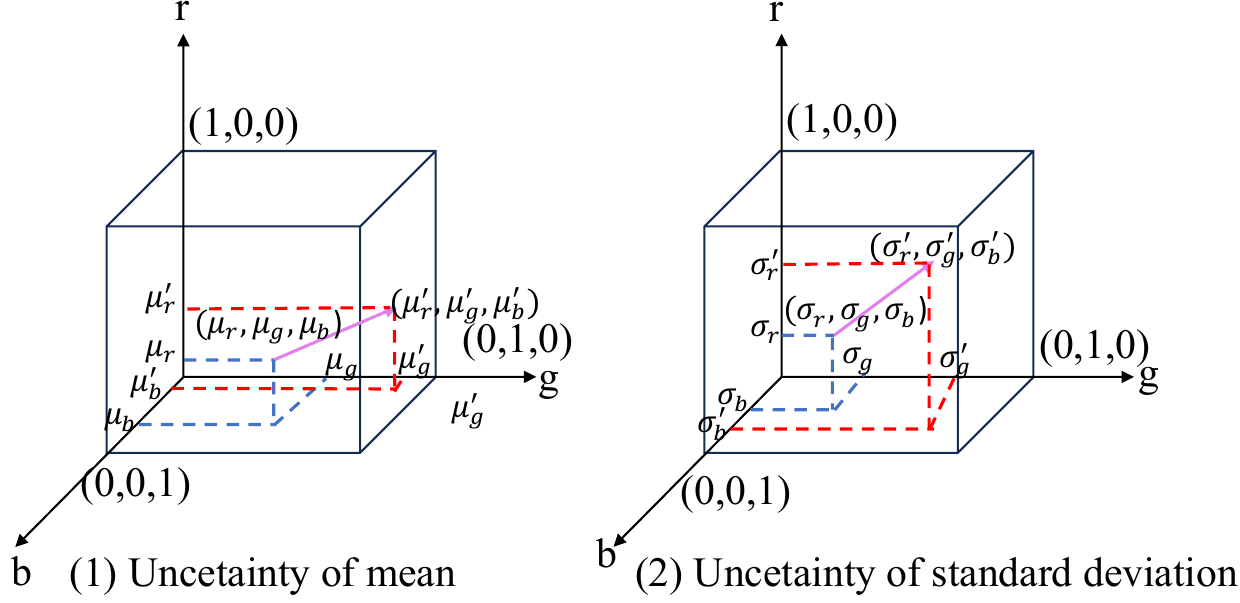}
\caption{Arrow refers to the direction of the perturbation, and the length is the perturbation amplitude.}
\label{fig:Uncetainty}
\end{figure}

To simulate more potential domains, it is necessary to account for the uncertainty of domain shifts, including the uncertainty in perturbation strength and direction. As illustrated in Figure ~\ref{fig:Uncetainty}, the projection of the line segment onto the RGB axes represents the perturbation magnitude for
each channel, while the arrow direction indicates the perturbation direction, jointly determined by the perturbation magnitudes of the three RGB channels. Motivated by the work~\cite{li2022uncertainty}, we use Gaussian distributions to model the uncertainty, assuming that perturbations are randomly sampled from the distribution. The distribution parameters are estimated as follows.

Given a batch of input images \(\mathit{x} \in \mathbb{R}^{B \times C \times H \times W}\), where  \(\mathit{H} \) represents image height, and \(\mathit{W} \) represents image width, \(\mathit{B} \) indicates batch-size, and \(\mathit{C=3} \) represents the number of channels, we compute the per-image channel-wise mean \(\mu_c(x)\) and standard deviation \(\sigma_c(x)\) over spatial dimensions for each channel \(c \in \{r, g, b\}\):

\begin{equation}
\mu_c(x) = \frac{1}{HW} \sum_{h,w} \mathit{x}_{c,h,w},
\end{equation}
\begin{equation}
\quad \sigma_c(x) = \sqrt{\frac{1}{HW} \sum_{h,w} (\mathit{x}_{c,h,w} - \mu_c(x)^2}.
\end{equation}

Then, we quantify the variations of these statistics \(\mu_c(x)\)  and \(\sigma_c(x)\) within a batch. They reflect inter-image appearance differences in the batch and also provide reasonable and meaningful range
for perturbations added to these statistics. The measure process is formulated as:

\begin{align}
\sigma_{\mu_c}^2(x) &= \frac{1}{B} \sum_{i=1}^B (\mu_c^i(x) - \bar{\mu}_c)^2,\\
\quad \sigma_{\sigma_c}^2(x) &= \frac{1}{B} \sum_{i=1}^B (\sigma_c^i(x) - \bar{\sigma}_c)^2.
\end{align}

where \(\bar{\mu}_c = \frac{1}{B} \sum_{i=1}^B \mu_c^i(x)\) and \(\bar{\sigma}_c = \frac{1}{B} \sum_{i=1}^B \sigma_c^i(x)\). These batch-level variations carry important style meaning:

\begin{itemize}
\item \(\sigma_{\mu_c}^2(x)\) reflects the degree of inter-image brightness or color tone variation in channel \(c\), indicating how much the mean intensity differs across images;
\item \(\sigma_{\sigma_c}^2(x)\) captures contrast or texture variation across the batch in that channel, measuring diversity in per-channel local detail.
\end{itemize}

In this paper, we suppose the perturbation to mean \(\mu_c(x)\) follow  \(\mathcal{N}(0, \sigma_{\mu_c}^2(x))\), and the perturbation to
variance \(\sigma_c(x)\) follow \(\mathcal{N}(0, \sigma_{\sigma_c}^2(x))\). By exploiting the given Gaussian distribution, random sampling can generate various new feature statistics information with different directions and intensities. The random perturbation can be formulated as:

\begin{align}
\mu_c'(x) &= \mu_c(x) + \epsilon_{\mu_c} \sigma_{\mu_c}(x), \quad \epsilon_{\mu_c} \sim \mathcal{N}(0, 1),\\
\sigma_c'(x) &= \sigma_c(x) + \epsilon_{\sigma_c} \sigma_{\sigma_c}(x), \quad \epsilon_{\sigma_c} \sim \mathcal{N}(0, 1).
\end{align}
where  \(\mu_c'(x)\) represents the mean after the perturbation, and \(\sigma_c'(x)\) represents the standard deviation after the perturbation.

The final augmented image \(\mathit{x}^* \)is computed as:

\begin{equation}
\mathit{x}^*_{c,h,w} = \frac{\mathit{x}_{c,h,w} - \mu_c(x)}{\sigma_c(x)} \sigma_c'(x) + \mu_c'(x),
\end{equation}

\begin{equation}
\mathit{x}^* = (\mathit{x}^*_{r,h,w}, \mathit{x}^*_{g,h,w}, \mathit{x}^*_{b,h,w}).
\end{equation}

\subsubsection{Structural Advantages}

\begin{itemize}
\item \textbf{No Structural Distortion}: As the operation only alters global per-channel statistics, local texture, edge, and geometric content remain unchanged.
\item \textbf{Training-only \& Plug-and-Play}: The module is only active during training and can be seamlessly inserted into any stereo backbone without architectural modifications or inference-time cost.
\item \textbf{Inter-sample Diversity Modeling}: Avoids collapsing all samples into a shared stylization regime.
\end{itemize}

\subsection{Feature Consistency Constraint}

Our data augmentation modifies only stylistic attributes of images (e.g., brightness, tone, and contrast) while preserving scene structure. If style variations in the input cause significant shifts in extracted matching features, the model may be overfitting to style-based shortcuts. To address this, we introduce a feature consistency constraint that encourages the model to extract shortcuts-invariant
feature representations.

We denote the original stereo pairs as \(\mathit{x_L}\) and \(\mathit{x_R}\), and their stylized variant generated by UgDA module as \(\mathit{x_L}^*\) and \(\mathit{x_R}^*\). Let \( f_{\text{feat}}(x_L) \), \( f_{\text{feat}}(x_R) \), \( f_{\text{feat}}(x_{L}^*) \) and \( f_{\text{feat}}(x_{R}^*) \) denote the output of the shared feature extractor. We define the feature consistency loss as:

\begin{equation}
\mathcal{L}_{\text{cons}} = \| f_{\text{feat}}(\mathit{x}_L) - f_{\text{feat}}(\mathit{x}_L^*) \|_2 + \| f_{\text{feat}}(\mathit{x}_R) - f_{\text{feat}}(\mathit{x}_R^*) \|_2,
\end{equation}

\begin{equation}
\mathcal{L} = \mathcal{L}_{\text{smooth}_{\mathit{L}_1}}(\hat{\mathit{d}}, \mathit{d}_{gt}) + \lambda \mathcal{L}_{\text{cons}}.
\label{eq:loss_total}
\end{equation}where $\hat{d}$ denotes the estimated disparity map, and $d_{gt}$ is the ground-truth disparity.

\(\mathcal{L}_{\text{cons}}\) penalizes feature-level deviations caused by  perturbations. Minimizing \(\mathcal{L}_{\text{cons}}\) encourages the model to rely on structural cues such as object boundaries rather than unstable visual shortcuts. This constraint is particularly important for\begin{table*}[t]
\centering
\small
\setlength{\tabcolsep}{2.5mm}
\renewcommand{\arraystretch}{0.95} 
\begin{tabular}{p{1cm}p{3cm}|cc|cc|cc|cc}
\hline
\multirow{2}{*}{\textbf{Base}} & \multirow{2}{*}{\textbf{Methods}} 
& \multicolumn{2}{c|}{\textbf{KITTI 2015}} 
& \multicolumn{2}{c|}{\textbf{KITTI 2012}} 
& \multicolumn{2}{c|}{\textbf{Middlebury(H)}} 
& \multicolumn{2}{c}{\textbf{ETH3D}} \\
 & & EPE & D1(3px) & EPE & D1(3px) & EPE & D1(2px) & EPE & D1(1px) \\
\hline
\multirow{3}{*}{--} 
& DSMNet & 1.46 & 6.5 & 1.26 & 6.2 & 2.62 & 13.8 & 0.69 & 6.2\\
& GANet & 2.31 & 11.7 & 1.93 & 10.1 & 5.41 & 20.3 & 1.33 & 14.1 \\
& CasStereo & 2.42 & 11.9 & 2.12 & 11.8 & 3.71 & 17.2 & 0.87 & 7.8 \\
\hline
\multirow{6}{*}{PSMNet} 
& PSMNet & 3.17 & 16.3 & 2.69 & 15.1 & 7.65 & 34.2 & 2.33 & 23.8 \\
& MS-PSMNet & 1.64 & 7.8 & 2.33 & 14.0 & 4.72 & 19.8 & 1.42 & 16.8 \\
& FC-PSMNet & 1.58 & 7.5 & 1.42 & 7.0 & 4.14 & 18.3 & 1.25 & 12.8 \\
& ITSA-PSMNet & 1.39 & 5.8 & 1.09 & 5.2 & 3.25 & 12.7 & 0.94 & 9.8 \\
& Graft-PSMNet & 1.32 & \textbf{5.3} & 1.09 & 5.0 & 2.34 & 10.9 & 1.16 & 10.7 \\
& \textbf{UgDA-PSMNet} & \textbf{1.19} & \textbf{5.3} & \textbf{1.01} & \textbf{4.8} & \textbf{2.30} & \textbf{8.5} & \textbf{0.64} & \textbf{10.2} \\
\hline
\multirow{4}{*}{GwcNet} 
& GwcNet & 3.43 & 22.7 & 2.77 & 20.2 & 7.23 & 37.9 & 2.78 & 54.2 \\
& FC-GwcNet & 1.72 & 8.0 & 1.45 & 7.4 & 5.14 & 21.1 & 1.13 & 11.7 \\
& ITSA-GwcNet & 1.33 & 5.4 & 1.02 & 4.9 & 2.73 & 11.4 & 0.62 & 7.1 \\
& \textbf{UgDA-GwcNet} &\textbf{1.13} & \textbf{4.9} & \textbf{0.92} & \textbf{4.2} & \textbf{1.82} & \textbf{8.3} & \textbf{0.38} & \textbf{5.7} \\
\hline
\multirow{3}{*}{CFNet} 
& CFNet & 1.71 & 6.0 & 1.04 & 5.2 & 3.24 & 15.4 & 0.48 & 5.7 \\
& ITSA-CFNet & \textbf{1.09} & \textbf{4.7} & \textbf{0.87} & \textbf{4.2} & 1.87 & 10.4 & 0.45 & 5.1 \\
& \textbf{UgDA-CFNet} & 1.13 & 5.2 & 0.90 & 4.7 & \textbf{1.80} & \textbf{8.2} & \textbf{0.39} & \textbf{4.9} \\
\hline
\end{tabular}
\caption{Performance comparison on multiple benchmarks (KITTI, Middlebury, ETH3D). Best results are in \textbf{bold}.}
\label{tab:comparison benchmarks}
\end{table*} SM, where the features of corresponding points should remain consistent regardless of style discrepancies between the left and right views. The total loss function is given by Equation\eqref{eq:loss_total}, where \(\lambda\) is a hyperparameter.

\section{Experiment}

\subsection{Dataset and Experimental Setting}

All models are trained on the SceneFlow dataset~\cite{mayer2016large}, which includes 35,454 training and 4,370 test stereo pairs with dense disparity maps.



To assess cross-domain generalization, models are tested on four real-world benchmarks with diverse domain characteristics:

\begin{itemize}
\item \textbf{KITTI 2012}~\cite{geiger2012we}: An outdoor dataset captured from real-world driving scenes, containing 194 stereo pairs with sparse disparity ground truth. The images are captured under varying lighting and object motion conditions, representative of daytime urban traffic environments.

\item \textbf{KITTI 2015} ~\cite{menze2015object}: An updated version of KITTI with 200 stereo pairs, featuring more dynamic scenes and additional moving objects. It provides sparse disparity annotations and high-resolution images (1242×375), and poses challenges in handling reflective surfaces, occlusion, and textureless areas.
\item \textbf{Middlebury} ~\cite{scharstein2014high}: This dataset contains 15 high-resolution stereo pairs (up to 5 megapixels), taken in controlled indoor environments with various lighting settings. It provides dense, pixel-accurate disparity maps. We use the half-resolution version for evaluation, as it is commonly adopted in domain generalization literature due to memory constraints.
\item \textbf{ETH3D} ~\cite{schops2017multi}: Consists of 27 grayscale stereo pairs captured in both indoor and outdoor scenes with varying lighting and object distances. The grayscale nature and lack of strong color cues make it a challenging benchmark for networks trained on RGB synthetic data.
\end{itemize}

We adopt End-Point Error (EPE) and D1 Error Rate as evaluation metrics, with thresholds of 3 px (KITTI), 2 px (Middlebury), and 1 px (ETH3D), respectively.


Experiments are conducted using PSMNet~\cite{chang2018pyramid}, GwcNet~\cite{guo2019group}, and CFNet~\cite{shen2021cfnet} as baselines. All models are implemented in PyTorch and trained for 20 epochs on a single RTX 4090 GPU with Adam optimizer (\(\beta_1 = 0.9\), \(\beta_2 = 0.999\)). The learning rate starts at 0.001 and decays to 0.0005 after 10 epochs. Batch size is set to 4, and the feature consistency loss weight $\lambda$ is 0.17.

\subsection{Performance Comparison}

Table~\ref{tab:comparison benchmarks} presents a comprehensive comparison of our UgDA-stereo models with three baselines and several representative methods including Graft-PSMNet~\cite{liu2022graftnet}, GANet~\cite{zhang2019ga}, CasStereo~\cite{gu2020cascade}, MS-PSMNet~\cite{cai2020matching}, ITSA~\cite{chuah2022itsa}, DSMNet~\cite{zhang2020domain}, FC-PSMNet, and FC-GwcNet~\cite{zhang2022revisiting}.

\begin{figure*}
    \centering
    \includegraphics[scale=0.66,keepaspectratio]{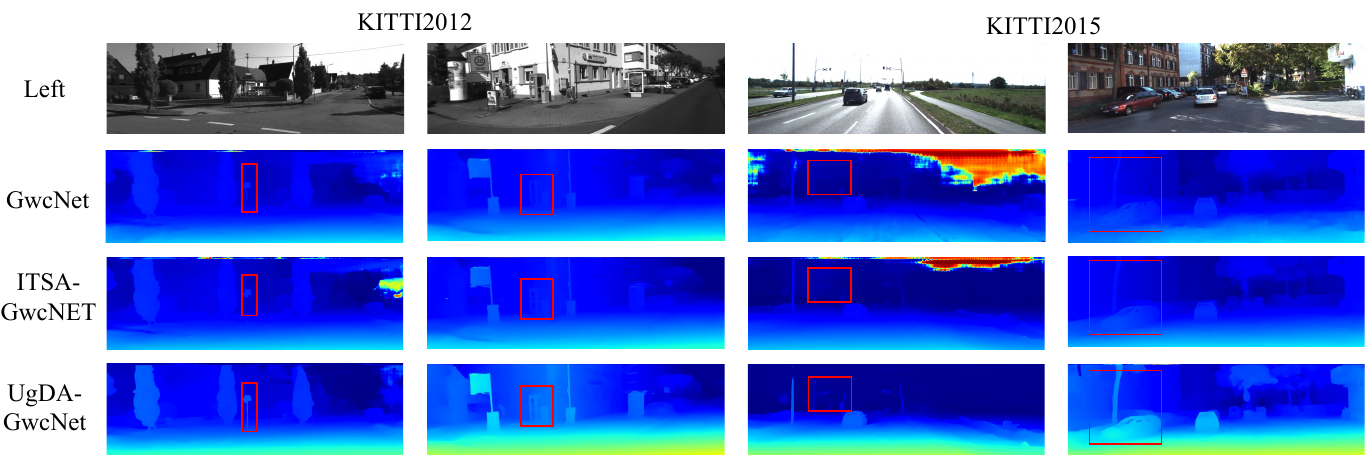}
    \caption{Visualization of the results of the baseline model GwcNet on KITTI2012 and KITTI2015}
    \label{fig:Visualization of the results}
\end{figure*}

UgDA-stereo consistently enhances the domain generalization performance of all baseline models. For example, compared to the original PSMNet, UgDA-PSMNet reduces D1 error by 10.3\%–25.7\% across KITTI2015, KITTI2012, Middlebury, and ETH3D. GwcNet achieves even greater improvements, with D1 error reductions of 16.0\%–48.46\% on the same datasets. The results demonstrate that our augmentation module improves model robustness under domain shifts, leading to more accurate and reliable disparity estimations in both indoor and outdoor scenes. Some visual results are shown in Figure~\ref{fig:Visualization of the results}.

\subsection{Learning Domain-Invariant Features}

\begin{figure}[t]
\centering
\includegraphics[width=\linewidth]{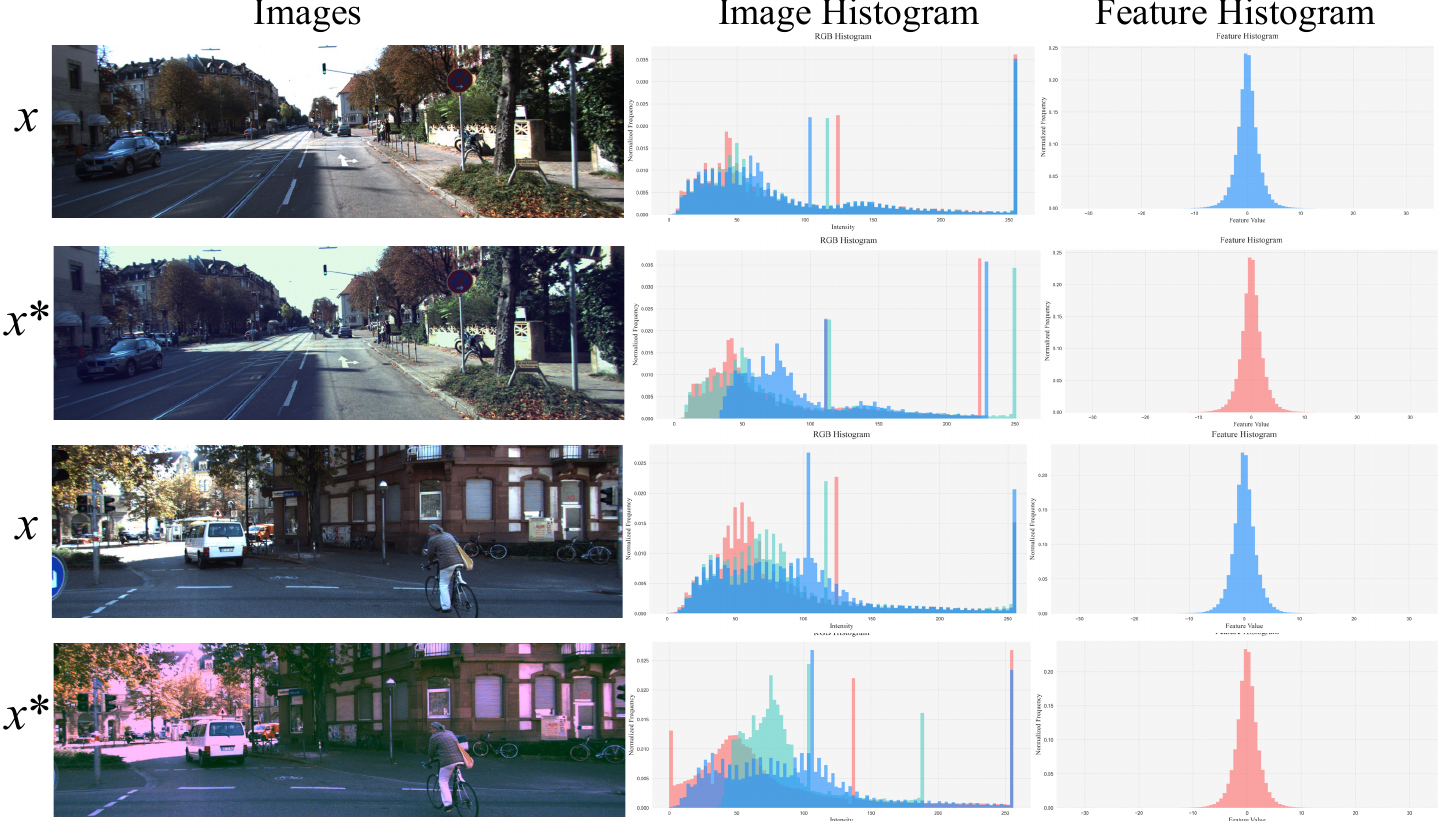}
\caption{Image and feature histogram comparison for original and augmented images.}
\label{fig:histogram}
\end{figure}

To better understand how UgDA-stereo affects model perception, we visualize both image histogram and intermediate feature histogram (averaged across channels) for perturbed and original images, as shown in Figure~\ref{fig:histogram}.

\begin{itemize}
\item UgDA-stereo leads to notable shifts in color tone and contrast, simulating domain shifts.
\item By applying the \(\mathcal{L}_{\text{cons}}\) constraint, the feature representations remain consistent before and after perturbation, demonstrating reduced reliance on shortcut cues.
\end{itemize}

\subsection{Ablation Studies}

We evaluate the contributions of our data augmentation (UgDA) module and feature consistency loss (\(\mathcal{L}_{\text{cons}}\)) through ablation studies on PSMNet and GwcNet, with results reported in Table~\ref{tab:ablation_gwcnet}. UgDA significantly reduces D1 error across KITTI 2012 and KITTI 2015. Adding \(\mathcal{L}_{\text{cons}}\) further enhances performance improvement. As shown in Table~\ref{tab:ablation_gwcnet}\begin{table}[t]
\centering
\small
\setlength{\tabcolsep}{0.8mm}
\renewcommand{\arraystretch}{0.9}
\label{tab:ablation}
\begin{tabular}{cc|cc|cc}
\toprule
\textbf{Augmentation} & \textbf{\(\mathcal{L}_{\text{cons}}\)} & \multicolumn{2}{c|}{\textbf{KITTI-2012}} & \multicolumn{2}{c}{\textbf{KITTI-2015}} \\
& & \textbf{PSMNet} & \textbf{GwcNet} & \textbf{PSMNet} & \textbf{GwcNet} \\
\midrule
\ding{55} & \ding{55} & 15.1 & 20.2 & 16.3 & 22.7 \\
\checkmark & \ding{55} & 5.8  & 4.9  & 6.1  & 5.9  \\
\checkmark & \checkmark & \textbf{4.8} & \textbf{4.2} & \textbf{5.3} & \textbf{4.9} \\
\bottomrule
\end{tabular}
\caption{Ablation study across multiple datasets based on D1 metric.}
\label{tab:ablation_gwcnet}
\end{table},  combined approach yields the lowest errors, highlighting the modules’ complementary roles in enabling generalizable stereo matching with minimal complexity.

\subsection{Robustness to Anomalous Scenarios}

To further evaluate the robustness of UgDA-stereo under different real-world domain shifts, we test our models on the DrivingStereo dataset, which contains images under various weather conditions including Sunny, Cloudy, Rainy, and Foggy. Table~\ref{tab:drivingstereo test} reports the D1 (3px) error across these conditions.

We highlight several key findings:

\noindent
 \includegraphics[width=\linewidth, height=0.7\linewidth]{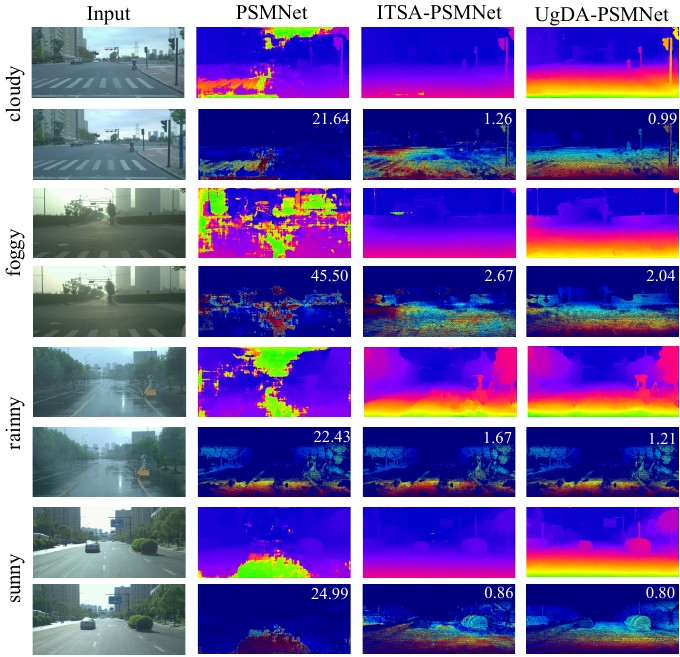}
\captionof{figure}{Qualitative results on the DrivingStereo dataset. For each group, top row shows the input images and the predicted disparity maps of PSMNet, ITSA-PSMNet, and our UgDA-PSMNet, respectively. Bottom row shows the EPE error maps of  different methods. The EPE values are marked in the upper right corner of error maps.}
\label{fig:Visualization of the DrivingStereo results}

\begin{itemize}
    \item \textbf{Robust Generalization Across Diverse Weather Conditions}: UgDA-enhanced models show consistent performance improvement across all four weather conditions. For example, UgDA-PSMNet achieves D1 errors of 4.2\% (Sunny), 3.1\% (Cloudy), 8.6\% (Rainy), and 5.3\% (Foggy), significantly outperforming all baseline methods. These results validate that our data augmentation mechanism effectively simulates real-world visual perturbations, enabling robust generalization under challenging visibility and lighting shifts.
    \item \textbf{Outperforms Fine-Tuned Baselines}: Notably, even without accessing any target-domain data during training, UgDA-stereo outperforms the fine-tuned versions of PSMNet and GwcNet on Rainy and Foggy subsets. This underscores the potential of domain-agnostic feature representation for real-world deployment.\begin{table}
\centering
\setlength{\tabcolsep}{1mm}
\renewcommand{\arraystretch}{1.0}
\begin{tabular}{p{2.5cm}ccccc}
\hline
\textbf{Methods} & \textbf{Sunny} & \textbf{Cloudy} & \textbf{Rainy} & \textbf{Foggy} & \textbf{Avg.} \\
\hline
PSMNet & 62.5 & 60.1 & 60.5 & 68.6 & 63.9 \\
FT-PSMNet & \textbf{4.0} & \textbf{2.9} & 11.5 & 6.5 & 6.3 \\
FC-PSMNet & 4.9 & 4.3 & 7.2 & 6.2 & 5.7 \\
ITSA-PSMNet & 4.8 & 3.2 & 9.4 & 6.3 & 5.9 \\
\textbf{UgDA-PSMNet} & 4.2 & 3.3 & \textbf{6.5} & \textbf{5.7} & \textbf{4.9} \\
\hline
GwcNet & 18.1 & 24.7 & 28.2 & 28.3 & 24.8 \\
FT-GwcNet & \textbf{3.1} & \textbf{2.5} & 12.3 & 6.0 & 6.0 \\
ITSA-GwcNet & 4.4 & 3.3 & 9.8 & 5.9 & 5.9 \\
\textbf{UgDA-GwcNet} & 3.7 & 3.4 & \textbf{6.2} & \textbf{5.5} & \textbf{4.7} \\
\hline
\end{tabular}
\caption{Robustness comparison of methods on DrivingStereo dataset in complex scenarios: \textit{Sunny}, \textit{Cloudy}, \textit{Rainy}, and \textit{Foggy}. D1 (3px) metric is used.}
\label{tab:drivingstereo test}
\end{table}
    \item \textbf{Visualization of the DrivingStereo results}: Figure~\ref{fig:Visualization of the DrivingStereo results} shows the qualitative comparison in different weather conditions, which further validates the potential of our method in enhancing the model robustness.
\end{itemize}

\section{Conclusion}

In this paper, we proposed UgDA-Stereo, a lightweight and effective data augmentation module designed to enhance domain generalization in SM. Our approach perturbs the per-channel mean and variance of RGB images using batch-driven uncertainty, simulating diverse appearance variations while maintaining geometric consistency. To further encourage shortcuts-invariant representations, we introduce a feature consistency loss that stabilizes feature extraction under style perturbations. The proposed method is simple, architecture-agnostic, and plug-and-play. Extensive experiments on real-world datasets show that UgDA-Stereo consistently boosts generalization performance across multiple baselines. Future work will explore advanced uncertainty modeling and strategies to address challenging regions such as occlusions and non-Lambertian surfaces.

\newpage  


\bibliography{Submission/main}

\makeatletter
\@ifundefined{isChecklistMainFile}{
  \newif\ifreproStandalone
  \reproStandalonetrue
}{
  \newif\ifreproStandalone
  \reproStandalonefalse
}
\makeatother

\ifreproStandalone
\documentclass[letterpaper]{article}
\usepackage[submission]{aaai2026}
\setlength{\pdfpagewidth}{8.5in}
\setlength{\pdfpageheight}{11in}
\usepackage{times}
\usepackage{helvet}
\usepackage{courier}
\usepackage{xcolor}
\frenchspacing

\begin{document}
\fi
\setlength{\leftmargini}{20pt}
\makeatletter\def\@listi{\leftmargin\leftmargini \topsep .5em \parsep .5em \itemsep .5em}
\def\@listii{\leftmargin\leftmarginii \labelwidth\leftmarginii \advance\labelwidth-\labelsep \topsep .4em \parsep .4em \itemsep .4em}
\def\@listiii{\leftmargin\leftmarginiii \labelwidth\leftmarginiii \advance\labelwidth-\labelsep \topsep .4em \parsep .4em \itemsep .4em}\makeatother

\setcounter{secnumdepth}{0}
\renewcommand\thesubsection{\arabic{subsection}}
\renewcommand\labelenumi{\thesubsection.\arabic{enumi}}

\newcounter{checksubsection}
\newcounter{checkitem}[checksubsection]

\newcommand{\checksubsection}[1]{%
  \refstepcounter{checksubsection}%
  \paragraph{\arabic{checksubsection}. #1}%
  \setcounter{checkitem}{0}%
}

\newcommand{\checkitem}{%
  \refstepcounter{checkitem}%
  \item[\arabic{checksubsection}.\arabic{checkitem}.]%
}
\newcommand{\question}[2]{\normalcolor\checkitem #1 #2 \color{blue}}
\newcommand{\ifyespoints}[1]{\makebox[0pt][l]{\hspace{-15pt}\normalcolor #1}}

\section*{Reproducibility Checklist}


\checksubsection{General Paper Structure}
\begin{itemize}

\question{Includes a conceptual outline and/or pseudocode description of AI methods introduced}{(yes/partial/no/NA)}
yes

\question{Clearly delineates statements that are opinions, hypothesis, and speculation from objective facts and results}{(yes/no)}
yes

\question{Provides well-marked pedagogical references for less-familiar readers to gain background necessary to replicate the paper}{(yes/no)}
yes

\end{itemize}
\checksubsection{Theoretical Contributions}
\begin{itemize}

\question{Does this paper make theoretical contributions?}{(yes/no)}
yes

	\ifyespoints{\vspace{1.2em}If yes, please address the following points:}
        \begin{itemize}
	
	\question{All assumptions and restrictions are stated clearly and formally}{(yes/partial/no)}
	yes

	\question{All novel claims are stated formally (e.g., in theorem statements)}{(yes/partial/no)}
	yes

	\question{Proofs of all novel claims are included}{(yes/partial/no)}
	yes

	\question{Proof sketches or intuitions are given for complex and/or novel results}{(yes/partial/no)}
	yes

	\question{Appropriate citations to theoretical tools used are given}{(yes/partial/no)}
	yes

	\question{All theoretical claims are demonstrated empirically to hold}{(yes/partial/no/NA)}
	yes

	\question{All experimental code used to eliminate or disprove claims is included}{(yes/no/NA)}
	yes
	
	\end{itemize}
\end{itemize}

\checksubsection{Dataset Usage}
\begin{itemize}

\question{Does this paper rely on one or more datasets?}{(yes/no)}
yes

\ifyespoints{If yes, please address the following points:}
\begin{itemize}

	\question{A motivation is given for why the experiments are conducted on the selected datasets}{(yes/partial/no/NA)}
	yes

	\question{All novel datasets introduced in this paper are included in a data appendix}{(yes/partial/no/NA)}
	NA

	\question{All novel datasets introduced in this paper will be made publicly available upon publication of the paper with a license that allows free usage for research purposes}{(yes/partial/no/NA)}
	NA

	\question{All datasets drawn from the existing literature (potentially including authors' own previously published work) are accompanied by appropriate citations}{(yes/no/NA)}
	yes

	\question{All datasets drawn from the existing literature (potentially including authors' own previously published work) are publicly available}{(yes/partial/no/NA)}
	yes

	\question{All datasets that are not publicly available are described in detail, with explanation why publicly available alternatives are not scientifically satisficing}{(yes/partial/no/NA)}
	NA

\end{itemize}
\end{itemize}

\checksubsection{Computational Experiments}
\begin{itemize}

\question{Does this paper include computational experiments?}{(yes/no)}
yes

\ifyespoints{If yes, please address the following points:}
\begin{itemize}

	\question{This paper states the number and range of values tried per (hyper-) parameter during development of the paper, along with the criterion used for selecting the final parameter setting}{(yes/partial/no/NA)}
	yes

	\question{Any code required for pre-processing data is included in the appendix}{(yes/partial/no)}
	yes

	\question{All source code required for conducting and analyzing the experiments is included in a code appendix}{(yes/partial/no)}
	yes

	\question{All source code required for conducting and analyzing the experiments will be made publicly available upon publication of the paper with a license that allows free usage for research purposes}{(yes/partial/no)}
	yes
        
	\question{All source code implementing new methods have comments detailing the implementation, with references to the paper where each step comes from}{(yes/partial/no)}
	yes

	\question{If an algorithm depends on randomness, then the method used for setting seeds is described in a way sufficient to allow replication of results}{(yes/partial/no/NA)}
	yes

	\question{This paper specifies the computing infrastructure used for running experiments (hardware and software), including GPU/CPU models; amount of memory; operating system; names and versions of relevant software libraries and frameworks}{(yes/partial/no)}
	yes

	\question{This paper formally describes evaluation metrics used and explains the motivation for choosing these metrics}{(yes/partial/no)}
	yes

	\question{This paper states the number of algorithm runs used to compute each reported result}{(yes/no)}
	yes

	\question{Analysis of experiments goes beyond single-dimensional summaries of performance (e.g., average; median) to include measures of variation, confidence, or other distributional information}{(yes/no)}
	yes

	\question{The significance of any improvement or decrease in performance is judged using appropriate statistical tests (e.g., Wilcoxon signed-rank)}{(yes/partial/no)}
	yes

	\question{This paper lists all final (hyper-)parameters used for each model/algorithm in the paper’s experiments}{(yes/partial/no/NA)}
	yes

\end{itemize}
\end{itemize}
\ifreproStandalone
\end{document}
\fi
\end{document}